\documentclass[runningheads,a4paper]{llncs}

\usepackage{amssymb}
\usepackage{graphicx}

\usepackage{xcolor}
\usepackage{mathtools}
\usepackage{amssymb}
\usepackage{booktabs}
\usepackage{comment}
\usepackage{xspace}

\newcommand*{\eg}{e.g.\@\xspace}
\newcommand*{\ie}{i.e.\@\xspace}

\usepackage{url}
\urldef{\mailsa}\path|{vedran.vukotic, christian.raymond, guillaume.gravier}@irisa.fr|
\urldef{\mailsb}\path|{s.l.pintea, j.c.vangemert}@tudelft.nl|    
\newcommand{\keywords}[1]{\par\addvspace\baselineskip
\noindent\keywordname\enspace\ignorespaces#1}

\begin{document}

\mainmatter  

\title{One-Step Time-Dependent\\Future Video Frame Prediction with a Convolutional Encoder-Decoder Neural Network}

\titlerunning{One-Step Time-Dependent Future Video Frame Prediction}

%
%
\author{Vedran Vukoti\'c \inst{1,2,3}
\and Silvia-Laura Pintea \inst{2}
\and Christian Raymond \inst{1,3}
\and \\ Guillaume Gravier \inst{1,4}
\and Jan C. van Gemert \inst{2}}
%

\institute{INRIA/IRISA Rennes, Rennes, France\\
\and TUDelft, Delft, The Netherlands\\
\and INSA Rennes, Rennes, France\\
\and CNRS, France\\
\mailsa\\
\mailsb}

%
%

\maketitle

\begin{abstract}
There is an inherent need for autonomous cars, drones, and other robots  to have a notion of how their environment behaves and to anticipate changes in the near future. In this work, we focus on anticipating future appearance given the current frame of a video.
Existing work focuses on either predicting the future appearance as the next frame of a video, or predicting future motion as optical flow or motion trajectories starting from a single video frame.  
This work stretches the ability of CNNs (Convolutional Neural Networks) to predict an anticipation of appearance at an arbitrarily given future time, not necessarily the next video frame. 
We condition our predicted future appearance on a continuous time variable that allows us to anticipate future frames at a given temporal distance, directly from the input video frame. 
We show that CNNs can learn an intrinsic representation of typical appearance changes over time and successfully generate realistic predictions at a deliberate time difference in the near future.
\end{abstract}


\section{Introduction}
For machines to successfully interact in the real world, anticipating actions and events and planning accordingly, is essential.
This is a difficult task, despite the recent advances in deep and reinforcement learning, due to the demand of large annotated datasets. 
If we limit our task to anticipating future appearance, annotations are not needed anymore. 
Therefore, machines have a slight advantage, as they can employ the vast collection of unlabeled videos available, which is perfectly suited for unsupervised learning methods. 
To anticipate future appearance based on current visual information, a machine needs to successfully be able to recognize entities and their parts, as well as to develop an internal representation of how movement happens with respect to time.

We make the observation that time is continuous, and thus, video frame-rate is an arbitrary discretization that depends on the camera sensor only. 
Instead of predicting the next discrete frame from a given input video frame, we aim at predicting a future frame at a given continuous temporal distance $\Delta t$ away from the current input frame. 
We achieve this by conditioning our video frame prediction on a time-related input variable.

In this work we explore one-step, long-term video frame prediction, from an input frame.
This is beneficial both in terms of computational efficiency, as well as avoiding the propagation and accumulation of prediction errors, as in the case of sequential\slash iterative prediction of each subsequent frame from the previous predicted frame.
Our work falls into the autoencoding category, where the current video frame is presented as input and an image resembling the anticipated future is provided as output.
Our proposed method consists of: an encoding CNN (Convolutional Neural Network), a decoding CNN, and a separate branch, parallel to the encoder, which models time and allows us to generate predictions at a given time distance in future. 

\subsection{Related Work}
\subsubsection{Predicting Future Actions and Motion.}
In the context of action prediction, it has been shown that it is possible to use high-level embeddings to anticipate future actions up to one second before they begin~\cite{forecastingActions}. 
Predicting the future event by retrieving similar videos and transferring this information, is proposed in \cite{yuen2010data}.
In \cite{lan2014hierarchical} a hierarchical representation is used for predicting future actions.
Predicting a future activity based on analyzing object trajectories is proposed in \cite{kitani2012activity}.
In \cite{huang2014action}, the authors forecast human interaction by relying on body-pose trajectories.
In the context of robotics, in \cite{koppula2016anticipating} human activities are anticipated by considering the object affordances.
While these methods focus on predicting high-level information --- the action that will be taken next, we focus on predicting low-level information, a future video frame appearance at a given future temporal displacement from a given input video frame. This has the added value that it requires less supervision.

Anticipating future movement in the spatial domain, as close as possible to the real movement, has also been previously considered. 
Here, the methods start from an input image at the current time stamp and predict motion --- optical flow or motion trajectories --- at the next frame of a video.
In \cite{liu2011sift} images are aligned to their nearest neighbour in a database and the motion prediction is obtained by 
transferring the motion from the nearest neighbor to the input image.
In~\cite{dejavu}, structured random forests are used to predict optical flow vectors at the next time stamp.
In~\cite{phaseCases}, the use of LSTM (Long Short Term Memory Networks) is advised towards predicting Eulerian future motion.
A custom deep convolutional neural network is proposed in~\cite{cnn2optiFlow} towards future optical flow prediction.
Rather than predicting the motion at the next video frame through optical flow, in \cite{variational} the authors propose to predict motion trajectories using variational autoencoders.
This is similar to predicting optical flow vectors, but given the temporal consistency of the trajectories, it offers greater accuracy. 
Dissimilar to these methods which predict future motion, we aim to predict the video appearance information at a given continuous future temporal displacement from an input video frame.

\subsubsection{Predicting Future Appearance.}
One intuitive trend towards predicting future information is predicting future appearance.
In \cite{walker2014patch}, the authors propose to predict both appearance and motion for street scenes using top cameras. 
Predicting patch-based future video appearance, is proposed in \cite{ranzato2014video}, by relying on large visual dictionaries.
In \cite{villegas2017learning} future video appearance is predicted in a hirarchical manner, by first predicting the video structure, and subsequently the individual frames. 
Similar to these methods, we also aim at predicting the appearance of future video frames, however we condition our prediction on a time parameter than allows us to perform the prediction efficiently, in one step.

Rather than predicting future appearance from input appearance information, hallucinating possible images has been a recent focus. 
The novel work in \cite{vondrick2016generating} relies on the GAN (Generative Adversarial Network) model \cite{gan} to 
create not only the appearance of an image, but also the possible future motion. 
This is done using spatio-temporal convolutions that discriminate between foreground and background.
Similarly, in \cite{tgan} a temporal generative neural network is proposed towards generating more robust videos. 
These generative models can be conditioned on certain information, to generate feasible outputs given the specific conditioning input~\cite{text2imageGAN}. 
Dissimilar to them, we rely on an autoencoding model.  
Autoencoding methods encode the current image in a representation space that is suitable for learning appearance and motion, and decode such representations to retrieve the anticipated future. 
Here, we propose to use video frame appearance towards predicting future video frames. 
However, we condition it on a given time indicator which allows us to predict future appearance at given temporal distances in the future. 

\section{Time-dependent Video Frame Prediction}
\label{method}
\begin{figure*}
    \centering
    \includegraphics[width=\textwidth]{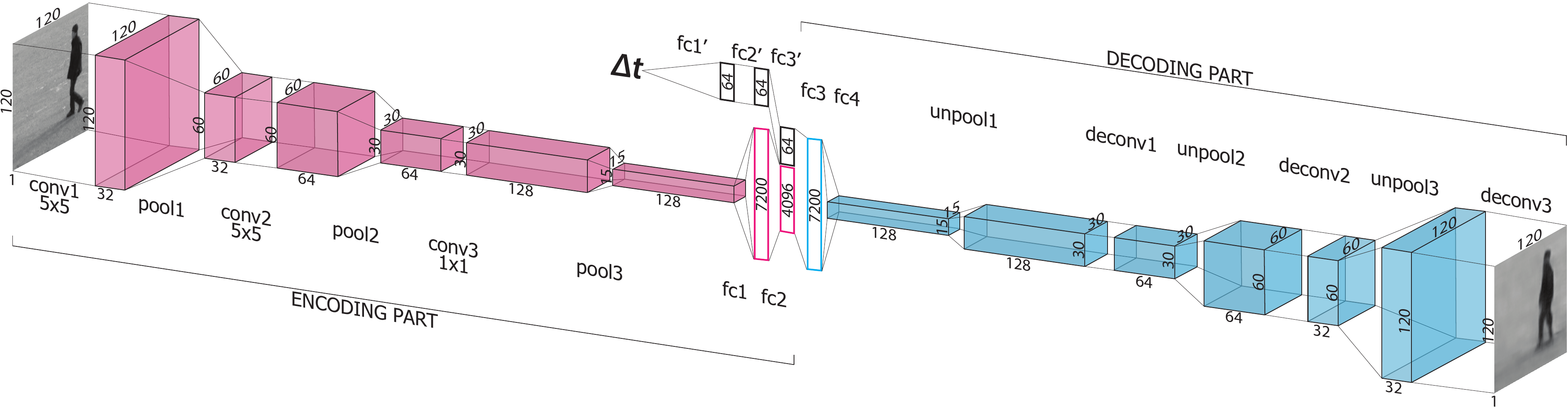}
	\caption{Our proposed architecture consists of two parts: i) an encoder part consisting of two branches: the first one taking the current image as input, and the second one taking as input an arbitrary time difference $\Delta t$ to the desired prediction and ii) a decoder part that generates an image, as anticipated, at the desired input time difference, $\Delta t$.}
    \label{fig:arch}
\end{figure*}

To tackle the problem of anticipating future appearance at arbitrary temporal distances, we deploy an encoder-decoder architecture. 
The encoder has two separate branches: one to receive the input image, and one to receive the desired temporal displacement $\Delta t$ of the prediction. 
The decoder takes the input from the encoder and generates a feasible prediction for the given input image and the desired temporal displacement.
This is illustrated in Figure~\ref{fig:arch}.
The network receives as inputs an image and a variable $\Delta t$, $\Delta t \in \mathbb{R}^+$, indicating the time difference from the time of the provided input image, $t_0$, to the time of the desired prediction.
The network predicts an image at the anticipated future time $t_0 + \Delta t$.
We use a similar architecture to the one proposed in~\cite{multi3d}. 
However, while their architecture is made to encode RGB images and a continuous angle variable to produce RGBD as output, 
our architecture is designed to take as input a monochromatic image and a continuous time variable, $\Delta t$, 
and to produce a monochromatic image, resembling a future frame, as output. 

More specifically, the architecture consists of the following:
\begin{enumerate}
\item \textit{an encoding part} composed of two branches:
\begin{itemize}
\item \textit{an image encoding branch} defined by 4 convolutional layers, 3 pooling layers and 2 fully-connected layers at the end; 
\item \textit{a time encoding branch} consisting of 3 fully-connected layers.
\end{itemize}
The final layers of the two branches are concatenated together, forming one bigger layer that is then provided to the decoding part.
\item \textit{a decoding part} composed of 2 fully-connected layers, 3 ``unpooling'' (upscaling) layers, and 3 ``deconvolutional'' (transpose convolutional) layers.
\end{enumerate}

The input time-indicator variable is continuous and allows for appearance anticipations at arbitrary time differences.
Training is performed by presenting to the network batches of $\{I_x, \Delta t, I_y\}$ tuples, where $I_x$ represents an input image at current relative time $t_0$, and $\Delta t$ represents a continuous variable indicating the time difference to the future video frame, and $I_y$ represents the actual video frame at $t_0 + \Delta t$. 

Predictions are obtained in one step. 
For every input image $I_x$ and continuous time difference variable $\Delta t$, a $\{I, \Delta t\}$ pair is given to the network as input, and an image representing the appearance anticipation $I_y$ after a time interval $\Delta t$ is directly obtained as output. No iterative steps are performed.

\section{Experiments}
\label{experiments}
\subsection{Experimental Setup}
We evaluate our method by generating images of anticipated future appearances at multiple time distances, and comparing them both visually and through MSE (Mean Squared Error) with the true future frames. 
We also compare to a CNN baseline that iteratively predicts the future video frame at $k \Delta t$ $(k=1, 2, ...)$ temporal displacements, from previous predictions. 

\subsubsection{Training parameters.}
During training, we use the Adam optimizer~\cite{adam}, with $L_2$ loss and dropout rate set to $80$\% for training. 
Training is performed up to 500,000 epochs with randomized minibatches consisting of 16 samples, where each sample contains one input image at current relative time $t_0=0$, a temporal displacement $\Delta t$ 
and the real target frame at the desired temporal displacement $\Delta t$. 
On a \textit{Titan X} GPU, training took approximately 16 hours with, on average, about 100,000 training samples (varying in each action category).
We argue that the type of action can be automatically detected, and is better incorporated by training a network per action category.
Thus, we opt to perform separate preliminary experiments for each action instead of training one heavy network to anticipate video frames corresponding to all the different possible actions.

\subsubsection{Network architecture.}
Given that the input, and thus also the output, image size is $120\times 120\times 1$ ($120\times 120$ grayscale images), in our encoder part, we stack convolutional and pooling layers that yield consecutive feature maps of the following decreasing sizes: $120\times 120$, $60\times 60$, $30\times 30$ and $15\times 15$, with an increasing number of feature maps per layer, namely 32, 64 and 128 respectively. 
Fully-connected layers of sizes 7,200 and 4,096 are added at the end. 
The separated branch of the encoder that models time consists of 4 fully connected layers of size 64, where the last layer is concatenated to the last fully-connected layer of the encoder convolutional neural network. 
This yields an embedding of size 4160 that is presented to the decoder. 
Kernel sizes used for the convolutional operations start at $5\times5$ in the first layers and decrease to $2\times2$ and $1\times1$ in the deeper layers of the encoder. 

For the decoder, the kernel sizes are the same as for the encoder, but ordered in the opposite direction.
The decoder consists of interchanging ``unpooling'' (upscaling) and ``deconvolutiton'' (transpose convolution) layers, 
yielding feature maps of the same sizes as the image-encoding branch of the encoder, only in the opposing direction. 
For simplicity, we implement pooling as a convolution with $2\times 2$ strips and unpooling as a 2D transpose convolution. 
\subsection{Dataset}
We use the KTH human action recognition dataset~\cite{kth} for evaluating our proposed method. 
The dataset consists of 6 different human actions, namely: \emph{walking}, \emph{jogging}, \emph{running}, \emph{hand-clapping}, \emph{hand-waving} and \emph{boxing}. 
Each action is performed by 25 actors. 
There are 4 video recordings for each action performed by each actor. 
Inside every video recording, the action is performed multiple times and information about the time when each action starts and ends is provided with the dataset.

To evaluate our proposed method, we randomly split the dataset by actors, in a training set --- with $80$\% of the actors, and a testing set --- with $20$\% of the actors. 
By doing so, we ensure that no actor is present in both the training and the testing split and that the network can generalize well with different looking people and does not overfit to specific appearance characteristics of specific actors. 
The dataset provides video segments of each motion in two directions --- \eg walking from right to left, and from left to right. 
This ensures a good setup for checking if the network is able to understand human poses and locations, and correctly anticipate the direction of movement. 
The dataset was preprocessed as follows: frames of original size $160 \times 120$ px were cropped to $120 \times 120$ px, and the starting\slash ending time of each action were adjusted accordingly to match the new cropped area. 
Time was estimated based on the video frame-rate and the respective frame number.

\subsection{Experimental Results}
\label{results}
\begin{figure}
    \centering
    \includegraphics[width=0.85\textwidth]{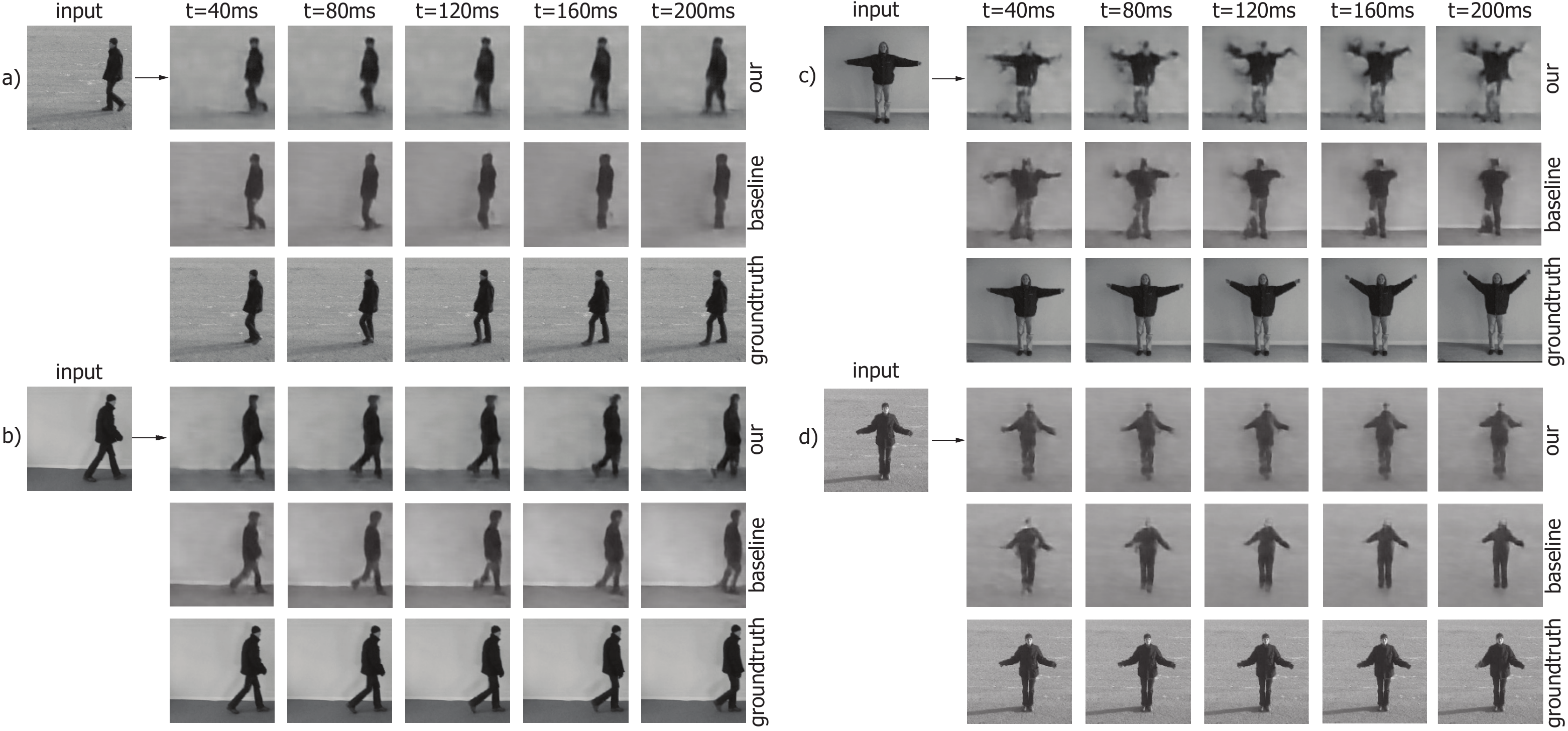}
	\caption {Comparison of predictions for a) a person walking to the left, b) a person walking to the right, c) a person waving their hands and d) a person slowly clapping with their hands. 
	The third set of images in each group represent the actual future frame --- the groundtruth.}
    \label{fig:results}
\end{figure}

Our method is evaluated as follows: an image at a considered time, $t_0=0$ and a time difference $\Delta t$ is given as input. 
The provided output represents the anticipated future frame at time $t_0+\Delta t$, where $\Delta t$ represents the number of milliseconds after the provided image.

The sequential encoder-decoder baseline method is evaluated by presenting solely an image, considered at time $t_0=0$ and expecting an image anticipating the future at $t_0 + \Delta t_b$ as output. 
This image is then fed back into the network in order to produce an anticipation of the future at time $t_0 + k \Delta t_b$, $k=1, 2, 3, ...$. 

\begin{figure}
    \centering
    \includegraphics[width=0.6\textwidth]{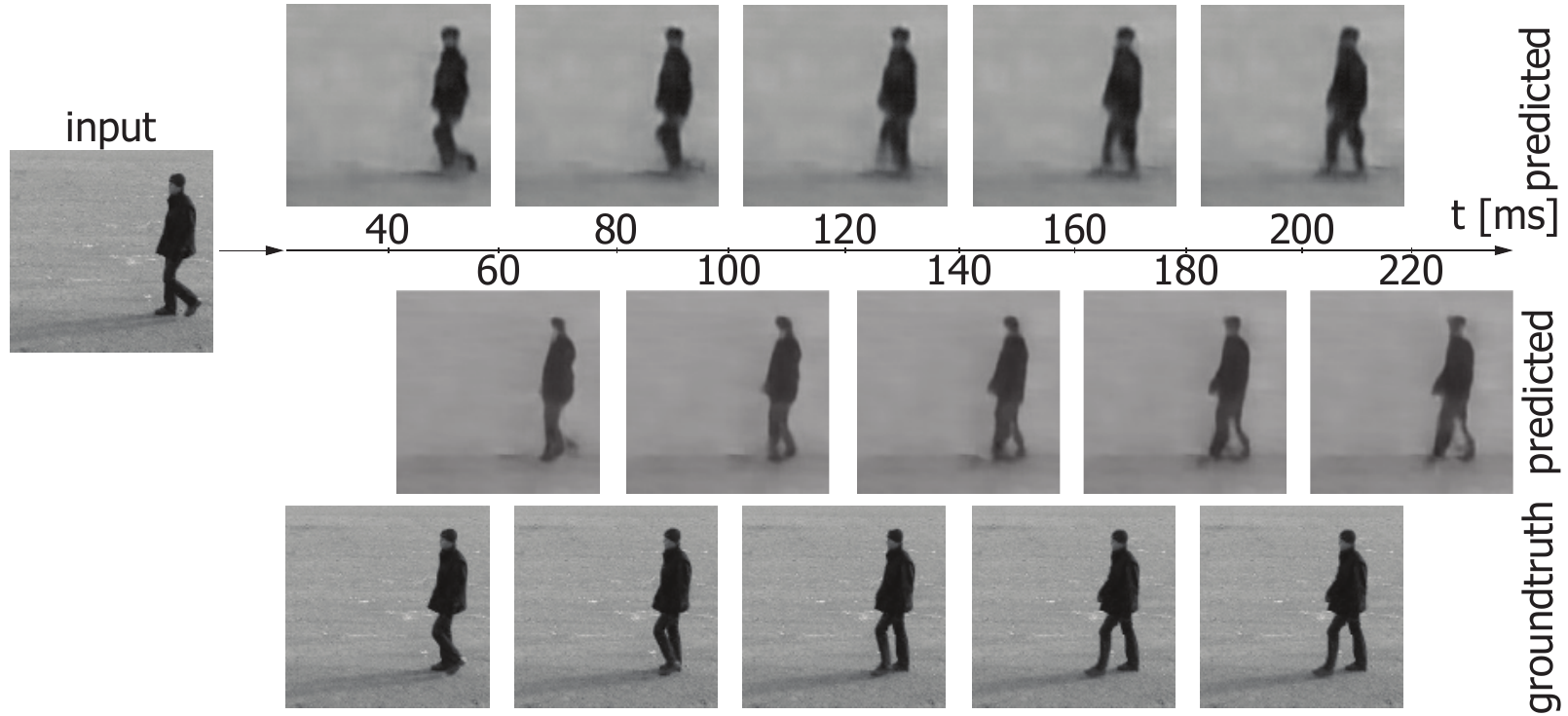}
	\caption{Prediction of seen and unseen temporal displacements.}
    \label{fig:unseen}
\end{figure}

For simplicity, we consider $t_0 = 0$ ms and refer to $\Delta t$ as simply $t$.
It is important to note that our method models time as a continuous variable. 
This enables the model to predict future appearances at previously unseen time intervals, as in Figure~\ref{fig:unseen}. The model is trained on temporal displacements defined by the framerate of the training videos. 
Due to the continuity of the temporal variable, it can successfully generate predictions for: i) temporal displacements found in the videos (\eg \textit{t=\{40 ms, 80 ms, 120 ms, 160 ms, 200 ms\}}), ii) unseen temporal displacement within the values found in the training videos (\eg \textit{t=\{60 ms, 100 ms, 140 ms, 180 ms\}}) and iii) unseen temporal displacement after the maximal value encountered during training (\eg \textit{t=220 ms}).

Figure ~\ref{fig:results}(a) illustrates a person moving from right to left, from the camera viewpoint, 
at walking speed. 
Despite the blurring, especially around the left leg when predicting for $t=120$ ms, our network correctly estimates the location of the person and position of body parts. 
Figure ~\ref{fig:results}(b) illustrates a person walking, from left to right. 
Our proposed network correctly localized the person and the body parts. 
The network is able to estimate the body pose, and thus the direction of movement and correctly predicts the displacement of the person to the right for any given time difference.  
The network captures the characteristics of the human gait, as it predicts correctly the alternation in the position of the legs.
The anticipated future frame is realistic but not always perfect, as it is hard to perfectly estimate walking velocity solely from one static image. 
This can be seen at $t=200$ ms in Figure~\ref{fig:results}(b). 
Our network predicts one leg further behind while the actor, as seen in the groundtruth, is moving slightly faster and has already moved their leg past the knee of the other leg.

\begin{figure*}
    \centering
    \includegraphics[width=\textwidth]{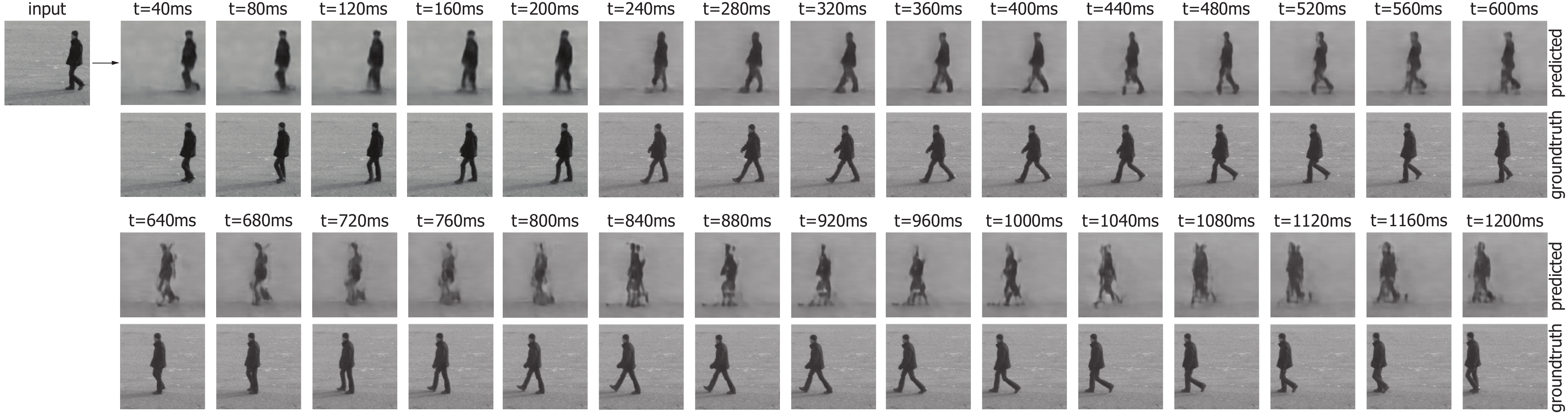}
	\caption {Long distance predictions. 
	For larger temporal displacements artifacting becomes visible. 
	The anticipated location of the person begins to differ from the groundtruth towards the end of the total motion duration.}
    \label{fig:long}
\end{figure*}

Our proposed network is able to learn an internal representation encoding the stance of the person such that it correctly predicts the location of the person, as well as anticipates their new body pose after a deliberate temporal displacement.
The baseline network does not have a notion of time and therefore relies on iterative predictions, which affects the performance. 
Figure~\ref{fig:results} shows that the baseline network loses the ability to correctly anticipate body movement after some time. 
Also in Figure~\ref{fig:results}(a) the baseline network correctly predicts the position of the legs up to $t=80$ ms, after that, it correctly predicts the global displacement of the person, but body part movements are not anticipated correctly. 
At $t>160$ ms the baseline network shows a large loss of details, enough to cause its inability to correctly model body movement. 
Therefore, it displays fused legs where they should be separated, as part of the next step the actor is making. 
Our proposed architecture correctly models both global person displacement and body pose, even at $t=200$ ms.

Figure~\ref{fig:results}(c) displays an actor \emph{handwaving}. 
Our proposed network successfully predicts upward movement of the arms and generates images accordingly. 
Here however, more artifacts are noticeable due to the bidirectional motion of hands during \emph{handwaving}, which is ambiguous.
It is important to note that although every future anticipation is independent from the others, they are all consistent: 
\ie it does not happen that the network predicts one movement for $t_1$ and a different movement for $t_2$ that is inconsistent with the $t_1$ prediction. 
This is a strong indicator that the network learns an embedding of appearance changes over time, the necessary filters relevant image areas and synthesizes correct future anticipations.

As expected, not every action is equally challenging for the proposed architecture. 
Table~\ref{mse} illustrate MSE scores averaged over multiple time differences, $t$, and for different predictions from the KTH test set. 
MSE scores were computed on dilated edges of the groundtruth images to only analyze the part around the person and remove the influence of accumulated variations of the background. A Canny edge detector was used on the groundtruth images. 
The edges were dilated by 11 pixels and used as a mask for both the groundtruth image and the predicted image. 
MSE values were computed solely on the masked areas.
We compare our proposed method with the baseline CNN architecture.
The average MSE scores, given in Table~\ref{mse}, show that our proposed method outperforms the encoder-decoder CNN baseline by a margin of $13.41$, on average, which is due to the iterative process of the baseline network.


\begin{table}
    \caption{Average MSE over multiple time distances and multiple video predictions, on the different action categories of KTH.
    We compare our method with the iterative baseline CNN, and show that our method on average performs better than the baseline in terms of MSE (lower is better).
    }
    \centering
    \begin{tabular}{lccccccc} \toprule

    \textbf{Method} & \textbf{Jogging} & \textbf{Running} & \textbf{Walking} & \textbf{Clapping} & \textbf{Waving} & \textbf{Boxing} & \textbf{Avg} \\ \midrule
    
    Baseline & 30.64 & 40.88 & 30.87 & 43.23 & 43.71 & 46.22 & 39.26 \\
    Our method & 11.66 & 17.35 & 19.26 & 33.93 & 35.19 & 37.71 & 25.85 \\ \bottomrule
    \end{tabular}
    \vspace{2mm}
    \label{mse}
\end{table}

\subsection{Ambiguities and Downsides}
\label{fails}
There are a few key factors that make prediction more difficult and cause either artifacts or loss of details in the predicted future frames. 
Here we analyze these factors.

\begin{figure} 
    \centering
    \includegraphics[width=0.5\textwidth]{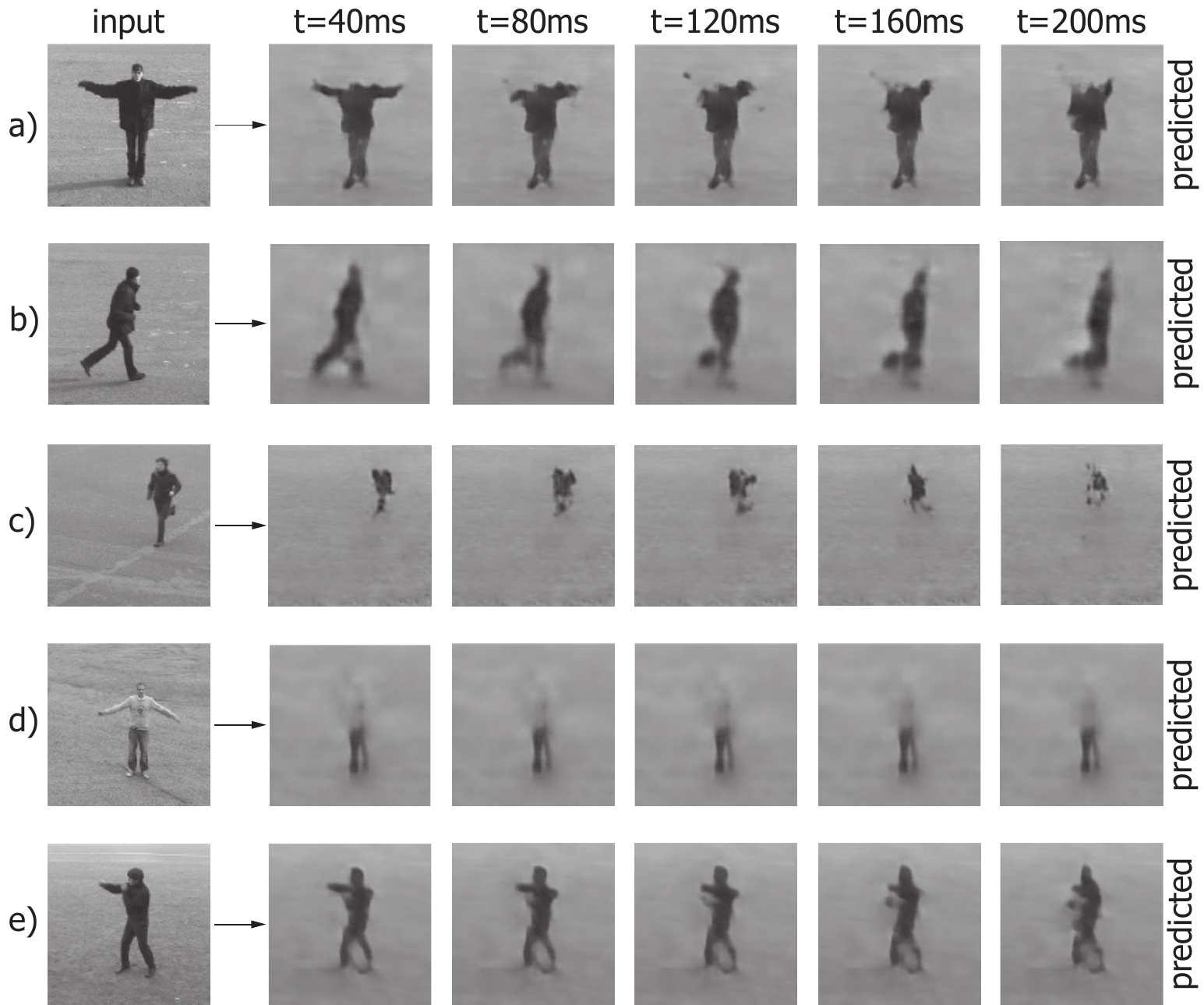}
	\caption {Examples of poorly performing future anticipations: a) loss of details in waving, b) loss of details in jogging, c) extreme loss of details in running, d) loss of details with low contrast and e) artifacts in boxing.}
    \label{fig:fails}
\end{figure}

\textbf{i) Ambiguities in body-pose} happen when the subject is in a pose that does contain inherent information about the future. 
A typical example would be when a person is waving, moving their arms up and down. 
If an image with the arms at a near horizontal position is fed to the network as input, this can results in small artifacts, as visible in Figure~\ref{fig:results}(c) where for larger time intervals $t$, there are visible artifacts that are part of a downward arm movement. 
A more extreme case is shown in Figure~\ref{fig:fails}(a) where not only does the network predict the movement wrong, but it also generates many artifacts with a significant loss of detail, which increases with the time difference, $t$.

\textbf{ii) Fast movement} causes loss of details when the videos provided for training do not offer a high-enough framerate. 
Examples of this can be seen in Figures~\ref{fig:fails}(b) and ~\ref{fig:fails}(c) where the increased speed in jogging and an even higher speed in running generate significant loss of details. 
Although our proposed architecture can generate predictions at arbitrary time intervals $t$, the network is still trained on discretized time intervals derived from the video framerate.
These may not be sufficient for the network to learn a good model. 
We believe this causes the loss of details and artifacts, and using higher framerate videos during training would alleviate this.

\textbf{iii) Decreased contrast} between the subject and the background describes a case where the intensity values corresponding to the subject are similar to the ones of the background.
This leads to an automatic decrease of MSE values, and a more difficult convergence of the network for such cases.
Thus, this causes to loss of details and artifacts.
This can be seen in Figure~\ref{fig:fails}(d). 
Such effect would be less prominent in the case in which color images would be used during training.

\textbf{iv) Excessive localization of movements} happens when the movements of the subject are small and localized. 
A typical example is provided by the boxing action, as present in the KTH dataset. 
Since the hand movement is close to the face and just the hand gets sporadically extended, the network has more difficulties in tackling this.
Despite the network predicting a feasible movement, often artifacts appear for bigger time intervals $t$, as visible in Figure~\ref{fig:fails}(e).

Despite the previously enumerated situations leading our proposed architecture to predictions that display loss of details and artifacts, most of these can be tackled and removed by either increasing the framerate, the resolution of the training videos, or using RGB information.

\section{Conclusion}
\vspace{-1.5mm}
In this work, we present a convolutional encoder-decoder architecture with a separate input branch that models time in a continuous manner. 
The aim is to provide anticipations of future video frames for arbitrary positive temporal displacements $\Delta t$, given a single image at current time $(t_0=0)$. 
We show that such an architecture can successfully learn time-dependant motion representations and synthesizes accurate anticipation of future appearance for arbitrary time differences $\Delta t > 0$. 
We compare our proposed architecture against a baseline consisting of an analogous convolutional encoder-decoder architecture that does not have a notion of time and relies on iterative predictions.
We show that out method outperforms the baseline both in terms of visual similarity to the groundtruth future video frames, as well as in terms of mean squared error with respect to it. 
We additionally analyze the drawbacks of our architecture and present possible solutions to tackle them. 
This work shows that convolutional neural networks can inherently model time without having a clear time domain representation. 
This is a novel notion that can be extended further and that generates high quality anticipations of future video frames for arbitrary temporal displacements. 
This is achieved without explicitly modelling the time period between the provided input video frame and the requested anticipation.




\end{document}